\documentclass[letterpaper, 10 pt, conference]{ieeeconf}  

\IEEEoverridecommandlockouts                              

\usepackage{graphicx} 
\usepackage{amsmath} 
\usepackage{amssymb}  
\usepackage{xcolor}
\usepackage{algorithm}
\usepackage{algorithmic}
\usepackage{subcaption}
\usepackage[font=small,labelfont=bf]{caption}

\title{\LARGE \bf Model-Based Quality-Diversity Search for Efficient Robot Learning}

\author{Leon Keller$^{1}$, Daniel Tanneberg$^{1}$, Svenja Stark$^{1}$, Jan Peters$^{1,2}$
\thanks{This project has received funding from the European Union's Horizon 2020 research and innovation programme under grant agreement No \#713010 (GOAL-Robots) and No \#640554 (SKILLS4ROBOTS).
Calculations for this research were conducted on the Lichtenberg high performance computer of the
TU Darmstadt.}
\thanks{$^{1}$ Intelligent Autonomous Systems, Technische Universit\"at Darmstadt}%
\thanks{$^{2}$ Robot Learning Group, Max-Planck Institute for Intelligent Systems}%
}

\begin{document}

\maketitle
\thispagestyle{empty}
\pagestyle{empty}

\begin{abstract}
Despite recent progress in robot learning, it still remains a challenge to program a robot to deal with open-ended object manipulation tasks.
One approach that was recently used to autonomously generate a repertoire of diverse skills is a novelty based Quality-Diversity~(QD) algorithm.
However, as most evolutionary algorithms, QD suffers from sample-inefficiency and, thus,  it is challenging to apply it in real-world scenarios.
This paper tackles this problem by integrating a neural network that predicts the behavior of the perturbed parameters into a novelty based QD algorithm. 
In the proposed Model-based Quality-Diversity search (M-QD), the network is trained concurrently to the repertoire and is used to avoid executing unpromising actions in the novelty search process. 
Furthermore, it is used to adapt the skills of the final repertoire in order to generalize the skills to different scenarios. 
Our experiments show that enhancing a QD algorithm with such a forward model improves the sample-efficiency and performance of the evolutionary process and the skill adaptation.
\end{abstract}

\section{Introduction}
Open-ended learning, or life-long learning, refers to a process where a robot has to autonomously acquire skills and knowledge by interacting with an environment \textit{forever}~\cite{thrun1995lifelong}.
In real-world scenarios robots are often exposed to changing environments or even need to solve new tasks without being trained on them beforehand, and not all future tasks can be foreseen to specify them for learning. 
As manually modifying the programming of the robot is time consuming and costly, it is a necessity that robots are able to deal with these challenges autonomously with little or no human intervention to develop throughout their lifespan~\cite{asada2009cognitive}.

A common task in life-long learning is to explore an unknown environment. Quality-diversity algorithms~\cite{qd} have been used to make robots autonomously explore a user-defined behavior space~\cite{expotocontrol}. 
These algorithms are typically evolutionary algorithms that aim at generating a repertoire of skills which is as diverse as possible by choosing the parents of each generation based on a novelty score. 
While these algorithms often are able to build a repertoire which covers the behavior space reasonably well, they suffer from sample-inefficiency -- a crucial limitation for robotic applications.
In this work, we address the sample-inefficiency problem by incorporating a neural network into the evolutionary process.
This model is trained from the samples in the repertoire and used to select the most promising children at each generation, thus avoiding to execute children that are unlikely to improve the repertoire. 
Quality-diversity search algorithms (with and without model) produce a discrete skill repertoire. 
Thus, in order to generalize to arbitrary desired behaviors in a continuous behavior space, skills in the repertoire need to be adapted. 
Our approach achieves this adaptation by using the gradient descent approach from~\cite{expotocontrol}, extended by the learned neural network.
The nearest neighbour of the desired behavior in the repertoire is selected and adapted to the desired behavior over multiple optimization steps. 

The contribution of this work is twofold: 
First, we propose a Model-based Quality-Diversity search (M-QD) algorithm that extendes QD search with a neural network that learns to score actions. 
Second, we propose an action adaptation strategy using this learned model. 
We show for both cases that adding the model helps to increase sample efficiency, and therefore reduces the number of expensive robot rollouts.
We evaluate the proposed M-QD and the adaptation with multiple tasks in a 2D environment and a robot environment.

\begin{figure}
    \centering
    \includegraphics[width=2.5in]{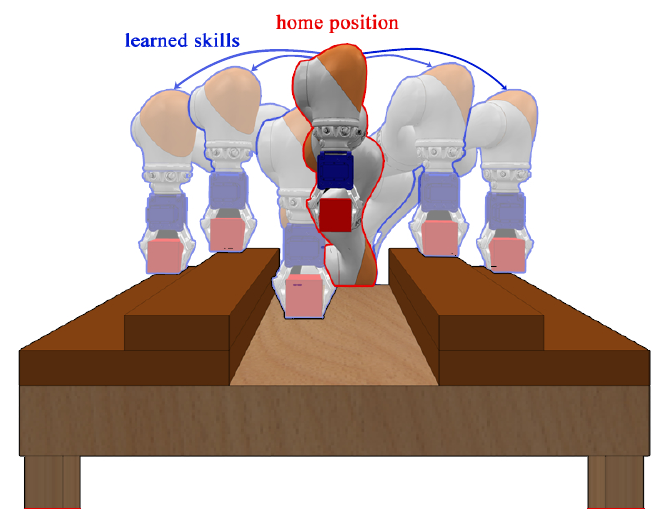}
    \caption{Overlay of four learned skills from the skill repertoire obtained by M-QD on the \texttt{Place} task, where the goal is to place the grasped objects safely onto the table and shelves.}
    \label{figure_intro}
    \vspace{-15pt}
\end{figure}

\subsection{Related Work}
A popular family of algorithms that has arisen from novelty search~\cite{novelty} is called Quality-Diversity search (QD)~\cite{qd} which has been applied to robotic tasks~\cite{expotocontrol}. 
These QD algorithms rate solutions based on a quality function and aim at finding qualitative solutions, in addition to building a diverse repertoire covering a defined behavior space as good as possible. 
The Novelty Search with Local Competition algorithm~(NSLC)~\cite{lc}, for example, tries to achieve these objectives by using a multi-objective genetic algorithm to optimize a nearest neighbour-based local quality function in addition to the novelty of found solutions. 
Behavioral repertoire evolution (BR-Evolution)~\cite{br}, in contrast, tracks all found solutions in an archive and progressively improves the quality of this archive by replacing solutions with better solutions that have a similar behavior. 
The multi-dimensional archive of phenotypic elites (MAP-elites)~\cite{map, map2} discretizes the behavior space in multiple bins and iteratively fills these bins with high quality solutions.

When using such novelty search algorithms to build a repertoire for a continuous behavior space, the skills of the discrete repertoire have to be generalized in order to estimate actions for arbitrary desired behaviors. 
One recent approach~\cite{expotocontrol} tries to achieve that by applying gradient descent, where the gradient of the mapping between actions and behaviors is estimated using a local linearization and samples from the repertoire. 
However, this approach often needs many adaptation steps and fails if the task is highly non-linear. 
Another approach uses a behavioral repertoire as a training set for a conditioned generative adversial network~\cite{gan} and then generates solutions for a given goal using this network. 
Similarly, the map-based Bayesian optimization algorithm (M-BOA)~\cite{map1} adapts a behavior-performance map, generated by MAP-elites, using Bayesian optimization. 

Models have been integrated into evolutionary search processes in a number of ways \cite{es}. Estimation of distribution algorithms \cite{es1} iteratively train a model to estimate the probability distribution of promising solutions and then, instead of mutation and crossover operators, use this model to generate new populations. Other algorithms directly try to build an inverse model of the mapping between objective and decision space \cite{es2, es3, es4}. Lastly, in a process called surrogate modelling \cite{es5, es6, es7, es8, es9}, many algorithms try to build a model of the objective function in order to reduce the number of computationally or otherwise expensive fitness evaluations. 

Similar to our method, the Surrogate-Assisted Illumination (SAIL) algorithm \cite{sail} aims to minimize the number of evaluations of a novelty search algorithm by integrating an approximated model of the objective function.

Our method is based upon a quality-diversity algorithm which iteratively builds an unstructured repertoire of skills~\cite{expotocontrol}. 
However, our approach is not purely evolutionary: similar as in model-based evolutionary learning, we incorporate a model into the QD algorithms to accelerate learning. 
Compared to popular model-based evolutionary learning, our method is most similar to surrogate modelling, however, our model predicts the behavior and quality of a given candidate instead of the fitness with respect to an objective function.
By using the predicted behavior and quality, we can eliminate samples that are unlikely to find a novel or to improve a known behavior, resulting in less samples that have to be evaluated on the robot.

\section{Learning A Repertoire of Diverse Skills}
As described in the introduction, our approach consists of two distinct parts.
The first part focuses on learning a diverse repertoire of skills covering a user-defined behavior space, whereas the second part focuses on adapting these skills in order to reach behaviors which are not present in the repertoire. 
We first introduce the quality-diversity search algorithm which builds the foundation of our approach, and introduce our model-based quality-diversity search variation. 
Lastly, we describe the gradient descent approach which is used to adapt the skills of the learned repertoires.

\subsection{Formalization}
We define an action $a \in A \subseteq \mathbb{R}^n$ as a vector of real valued parameters. 
When an action gets executed, the corresponding environment transitions from a start state $s_{\text{start}} \in S \subseteq \mathbb{R}^n$ to a destination state $s_{\text{dest}} \in S \subseteq \mathbb{R}^n$. 
We refer to this destination state as the behavior of an action. 
This behavior $s_{\text{dest}} = b$ is task-specific and can include any observable state of the robot (e.g. joint angles) and/or its environment (e.g. position of objects). 
Additionally, each action gets assigned a task-specific quality score $q$. 
Actions are evaluated with respect to their behavior $b$ and their quality $q$ using the evaluation functions
$$
f_b(a) = b \ \text{ and } f_q(a) = q \ .
$$
We refer to the tuple $s = (a, b, q)$ as a skill, including its action $a$, and the corresponding behavior $b$ and quality $q$.

\subsection{Quality-Diversity Search}
Quality-Diversity (QD) search is a novelty-based evolutionary search algorithm. 
In contrast to classical evolutionary algorithms, the evolution is mainly driven by novelty, which measures how unique the behavior of an skill is, instead of a fitness function. 
Additionally, QD uses a task-specific quality score to prefer better performing actions whenever multiple actions produce a similar behavior.
The goal of a QD algorithm is to learn a repertoire of diverse and good performing skills, covering a given or learned behavior space.

The novelty of skills in the repertoire is measured according to a novelty score function. 
In this work, we use the average Euclidean distance to the $K$ nearest neighbours as novelty score function, similar as in~\cite{expotocontrol,novelty} and given by
$$
\text{nov}(a) = \frac{1}{K} \sum_{i=1}^K d(f_b(a_{nn_i}), f_b(a)) \ ,
$$
where $d(.)$ denotes the Euclidean distance and $\{a_{nn_1}, \dots, a_{nn_K}\}$ are the $K$ nearest neighbours of $a$, measured in behavior space. More precisely, the nearest neighbours are given by the $K$ actions in the repertoire whose behavior is closest to the behavior of $a$, measured using the Euclidean distance. 

The algorithm is initialized with an empty repertoire. 
The first generation of skills is created randomly by sampling uniformly in action space.
All following generations are created by applying crossover and mutation operators to parent skills, which are sampled from the repertoire  proportionally to their novelty score. 
Every time a new generation is created, all its actions get evaluated in order to get their behavior and quality. 
After that, every candidate skill is compared to its nearest neighbour in the current repertoire. 
If the distance between the behavior and its closest neighbour is greater than a predefined threshold value $t_{\text{dist}}$, the new skill is added to the repertoire.
If the distance is smaller than the threshold $t_{\text{dist}}$, the new skill replaces the closest neighbour, if it performs better according to the task-specific quality score. 
Otherwise, the skill is not added to the repertoire. 
All skills that are removed or not added to the repertoire are saved in a separate archive. 
After all skills of a generation are processed, the novelty score of all skills in the repertoire is updated and a new generation is formed.
This process is repeated for a number of generations in order to form a large, diverse and highly qualitative repertoire.
\begin{algorithm}[t]
\caption{Model-based quality-diversity search (M-QD)}
\label{mqd_code}

\begin{algorithmic}
\begin{small}
\STATE Initialize empty repertoire 
\STATE Initialize model $\phi$ with random parameters 

\WHILE{current \ generation $\leq$ max \ generation}
\STATE parents $\leftarrow$ repertoire.sample(population size)
\STATE population $\leftarrow$ populate(parents)
\STATE survivor $\leftarrow \emptyset$

\FORALL{action a $\in$ population}
\STATE $\Tilde{b}$, $\Tilde{q}$ $\leftarrow \phi$(a)
\IF{$\widetilde{\text{nov}}(a, \Tilde{b}) > t_{\text{nov}}$ or $\widetilde{\text{qua}}(a, \Tilde{q}) > t_{\text{qua}}$}
\STATE b, q $\leftarrow$ evaluate(a)
\STATE $b_{nn}, q_{nn} \leftarrow$ repertoire.closest(b)
\IF{$d(b, b_{nn}) > t_{\text{dist}}$}
\STATE repertoire.add((a, b, q))
\ELSIF{$q > q_{nn}$}
\STATE repertoire.replace(nn, (a, b, q))
\ENDIF
\ENDIF
\ENDFOR

\STATE update novelty score of all skills in the repertoire
\STATE update model $\phi$ using mini-batches from the repertoire
\ENDWHILE

\end{small}
\end{algorithmic}
\end{algorithm}


\begin{figure*}[ht]
\begin{subfigure}{.2\textwidth}
  \centering
  \includegraphics[width=.8\linewidth]{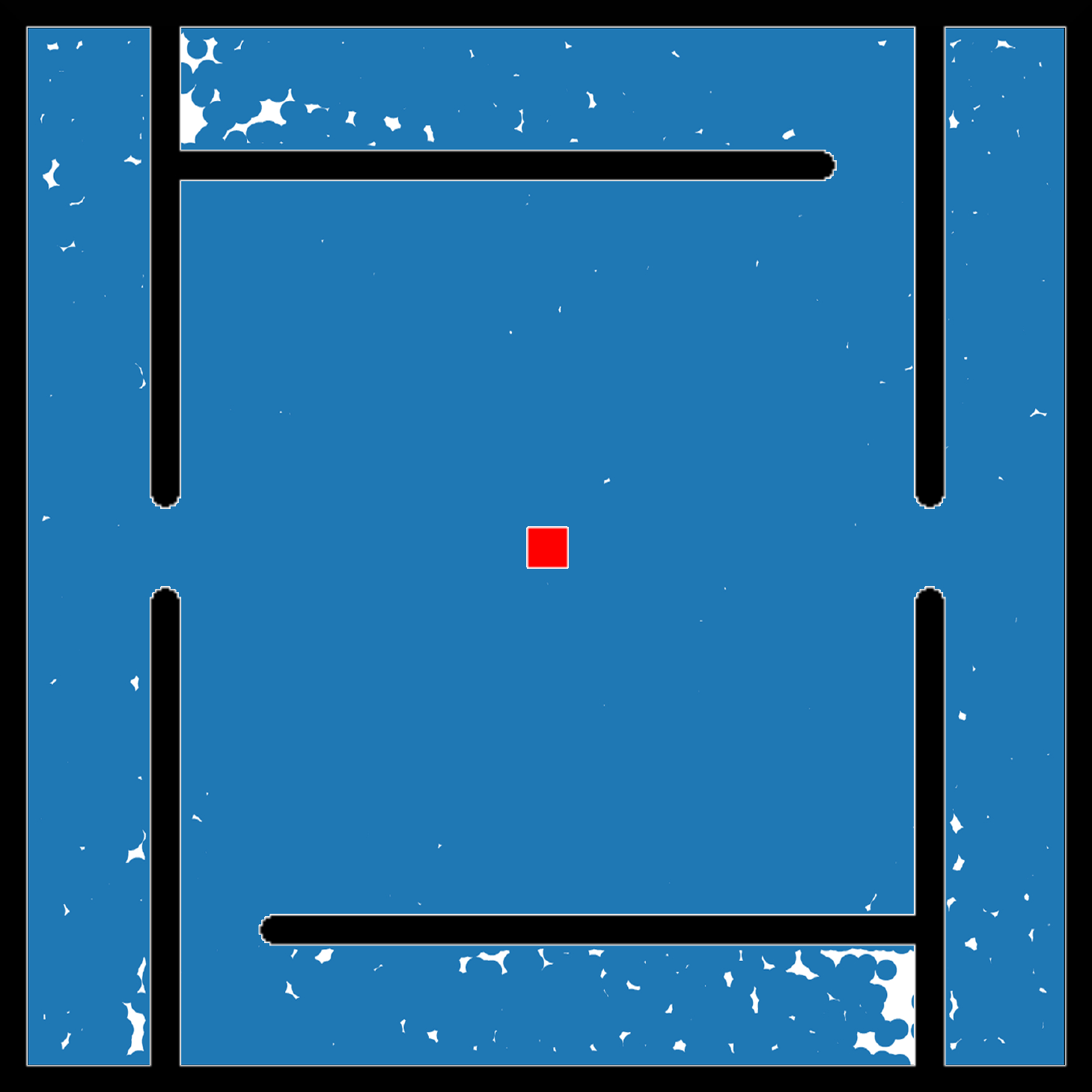}  
\end{subfigure}%
\begin{subfigure}{.2\textwidth}
  \centering
  \includegraphics[width=.8\linewidth]{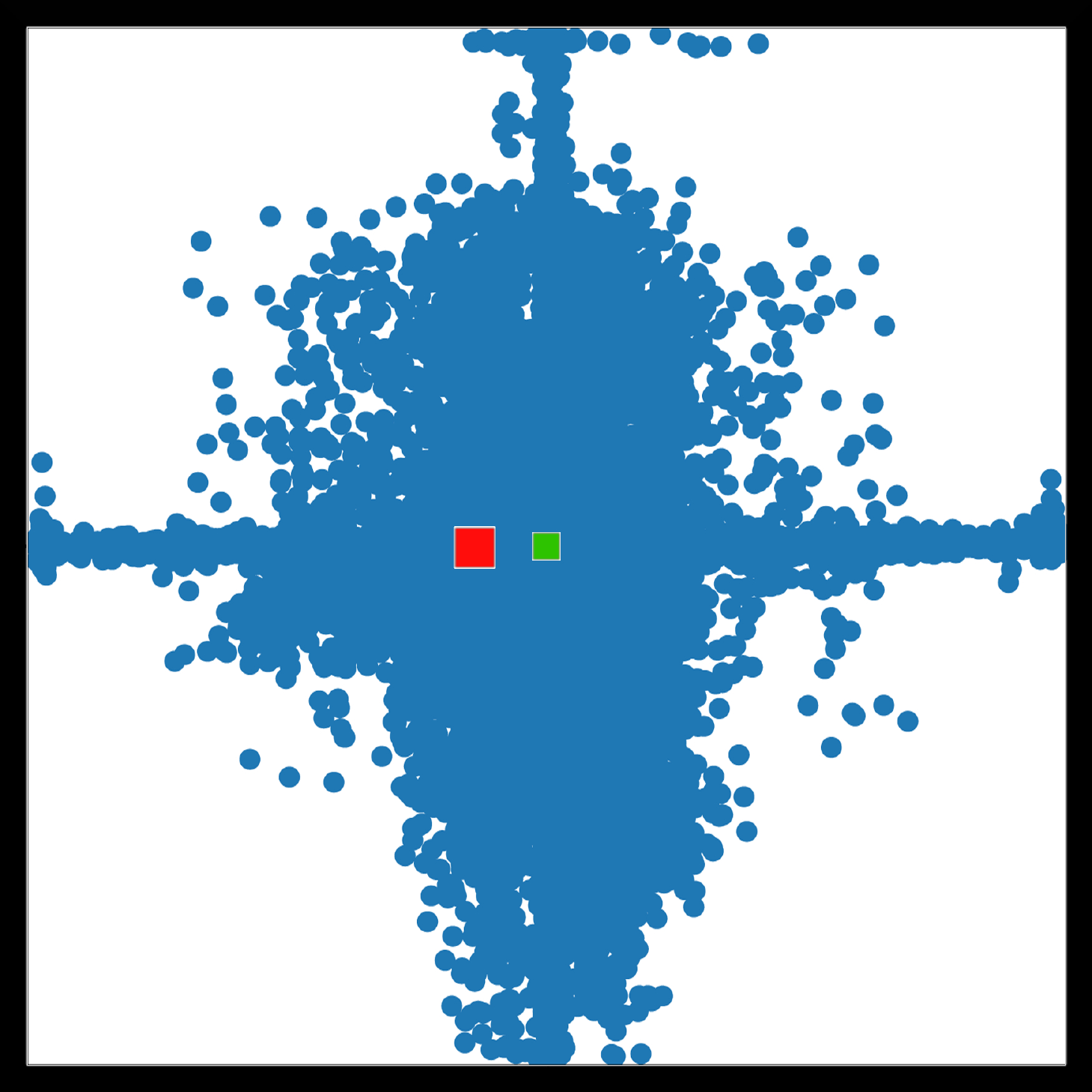}  
\end{subfigure}%
\begin{subfigure}{.2\textwidth}
  \centering
  \includegraphics[width=.8\linewidth]{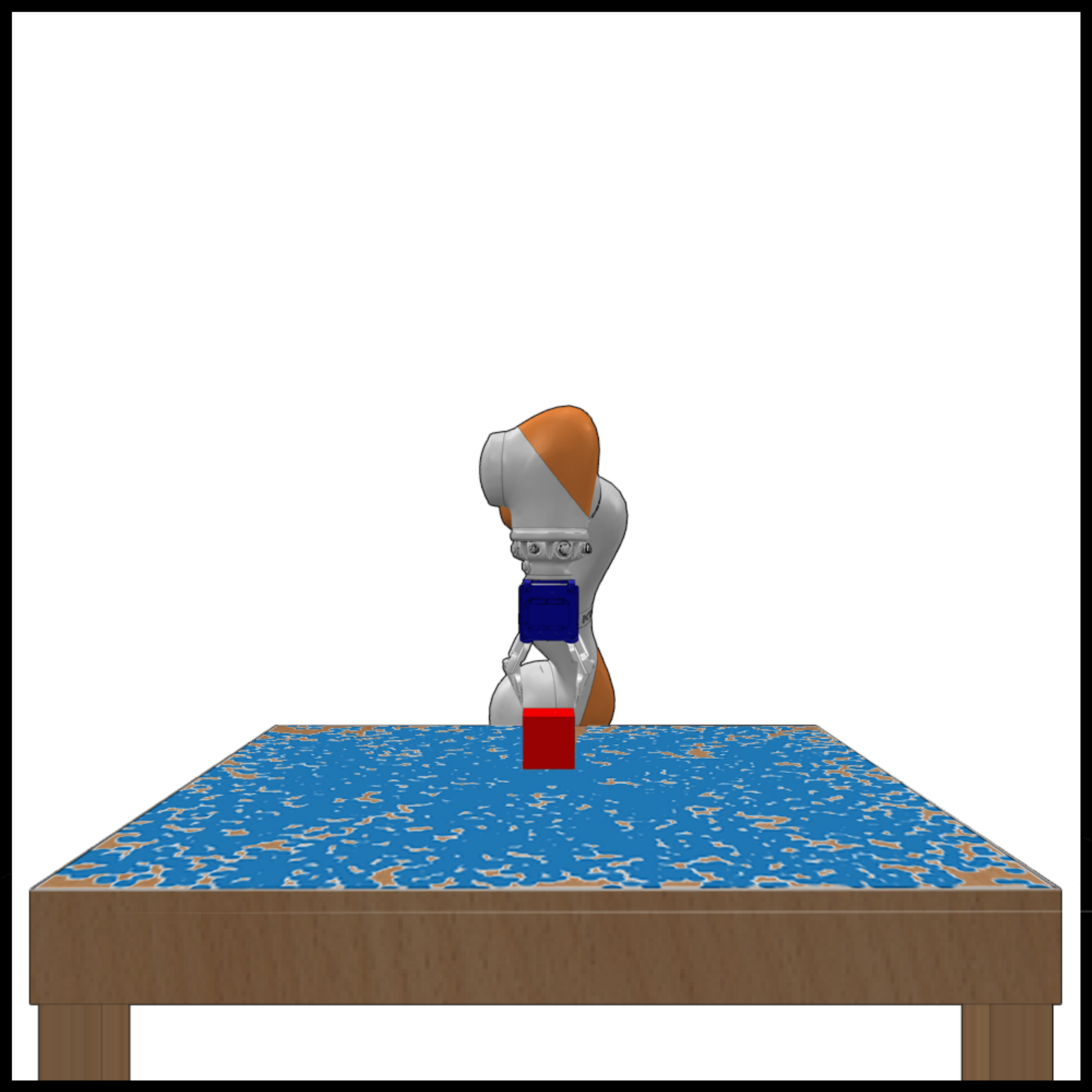}  
\end{subfigure}%
\begin{subfigure}{.2\textwidth}
  \centering
  \includegraphics[width=.8\linewidth]{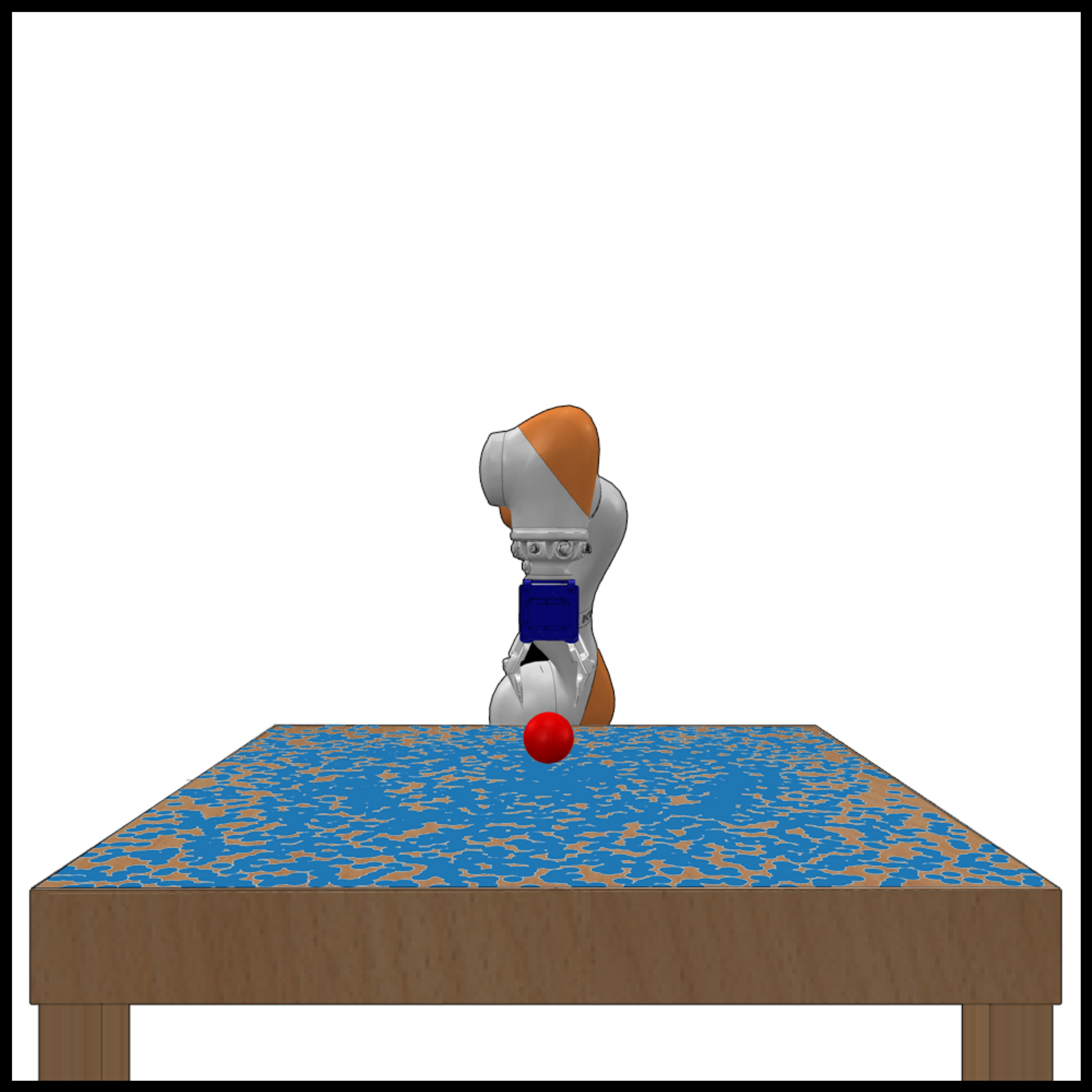}  
\end{subfigure}%
\begin{subfigure}{.2\textwidth}
  \centering
  \includegraphics[width=.8\linewidth]{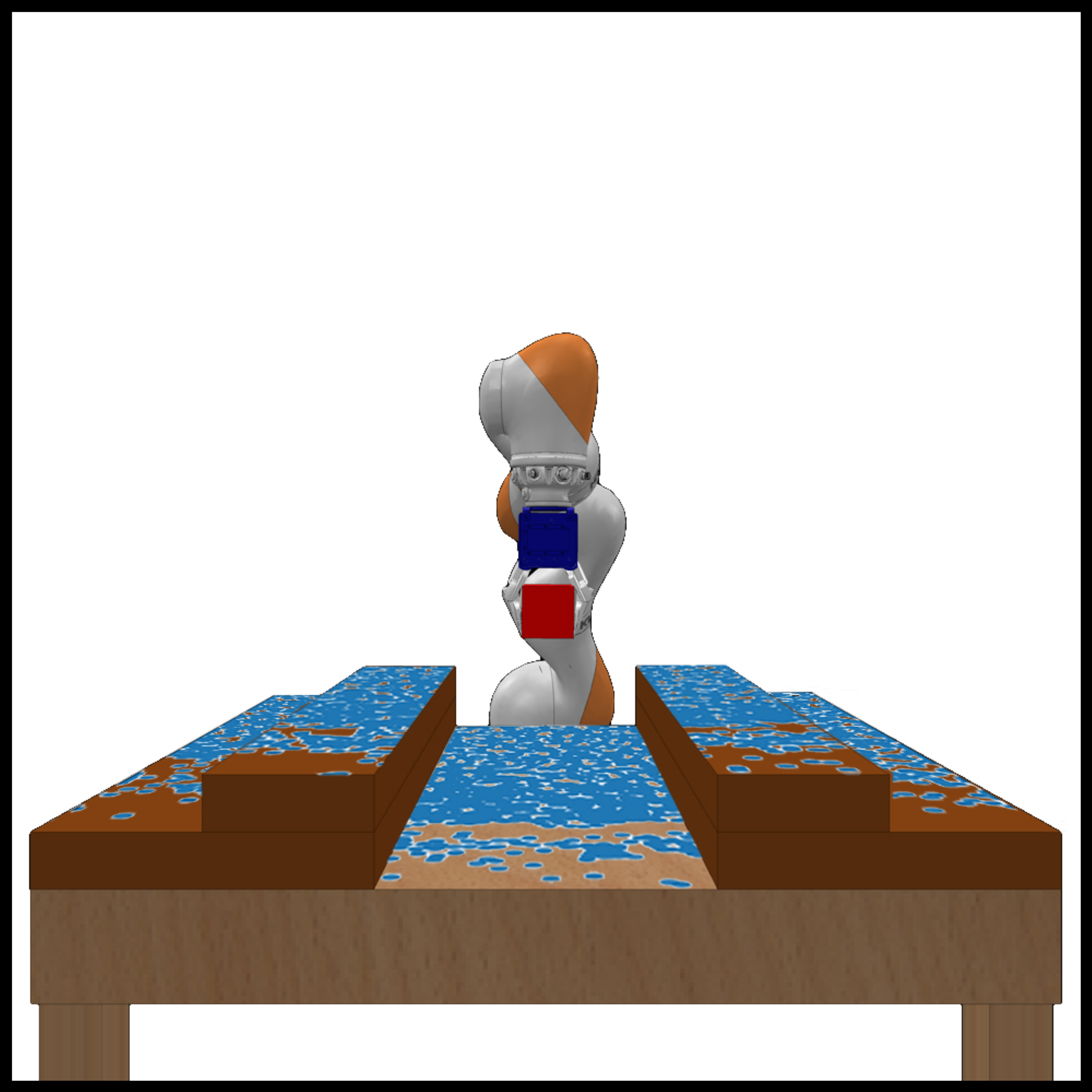}  
\end{subfigure}


\begin{subfigure}{.2\textwidth}
  \centering
  \includegraphics[width=.9\linewidth]{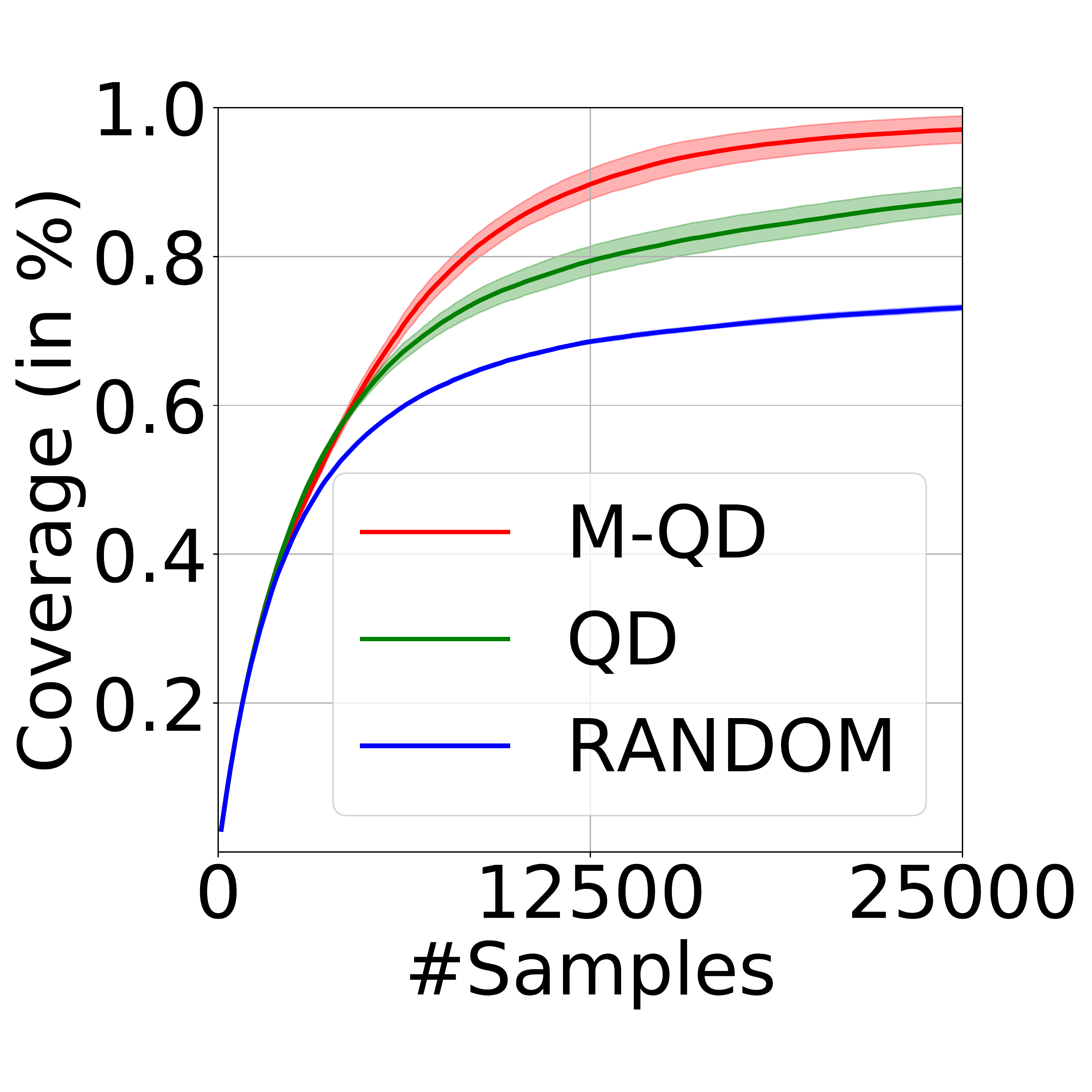}  
  \caption{\texttt{2D Obstacle}}
  \label{figure_toytask1}
\end{subfigure}%
\begin{subfigure}{.2\textwidth}
  \centering
  \includegraphics[width=.9\linewidth]{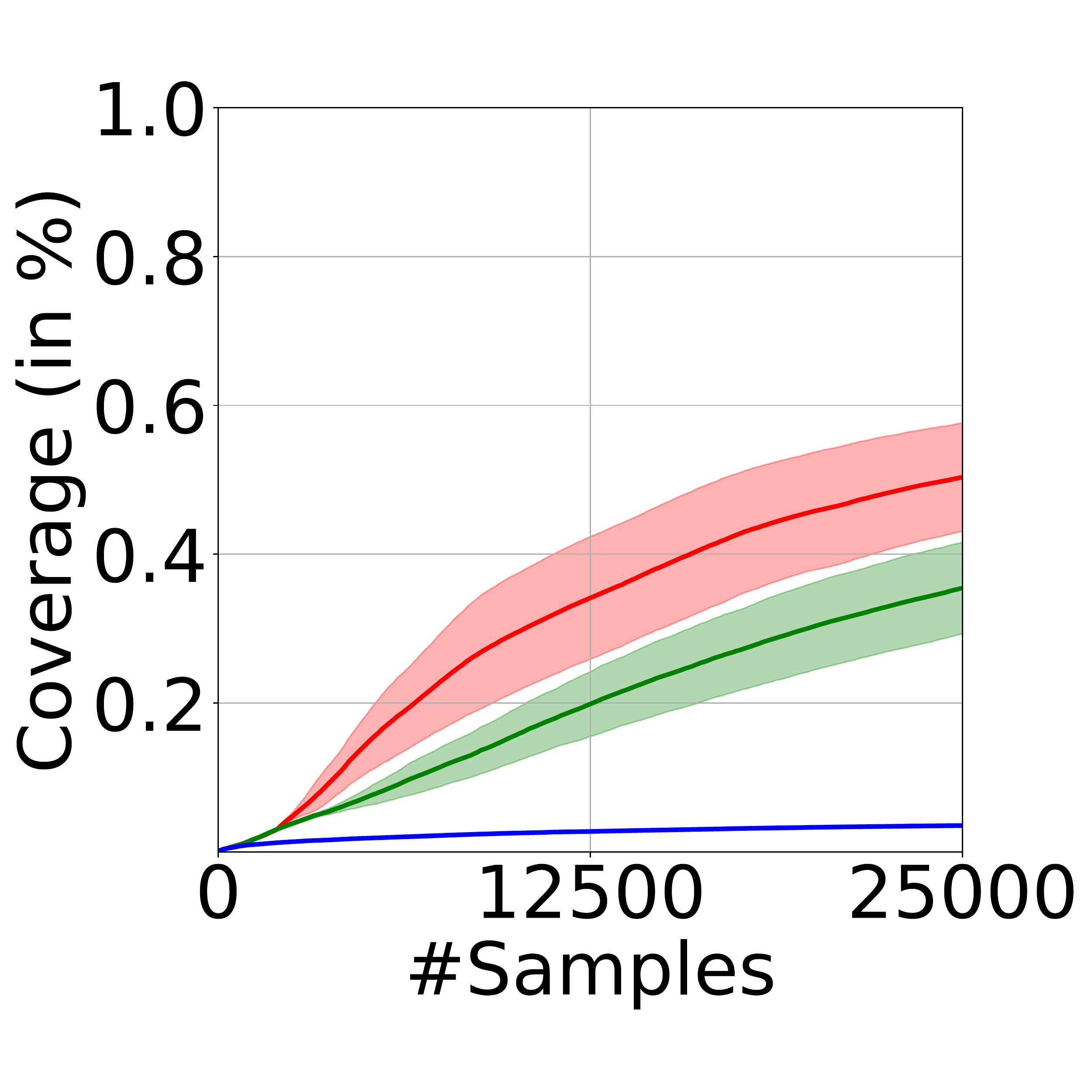}  
  \caption{\texttt{2D Object}}
  \label{figure_toytask2}
\end{subfigure}%
\begin{subfigure}{.2\textwidth}
  \centering
  \includegraphics[width=.9\linewidth]{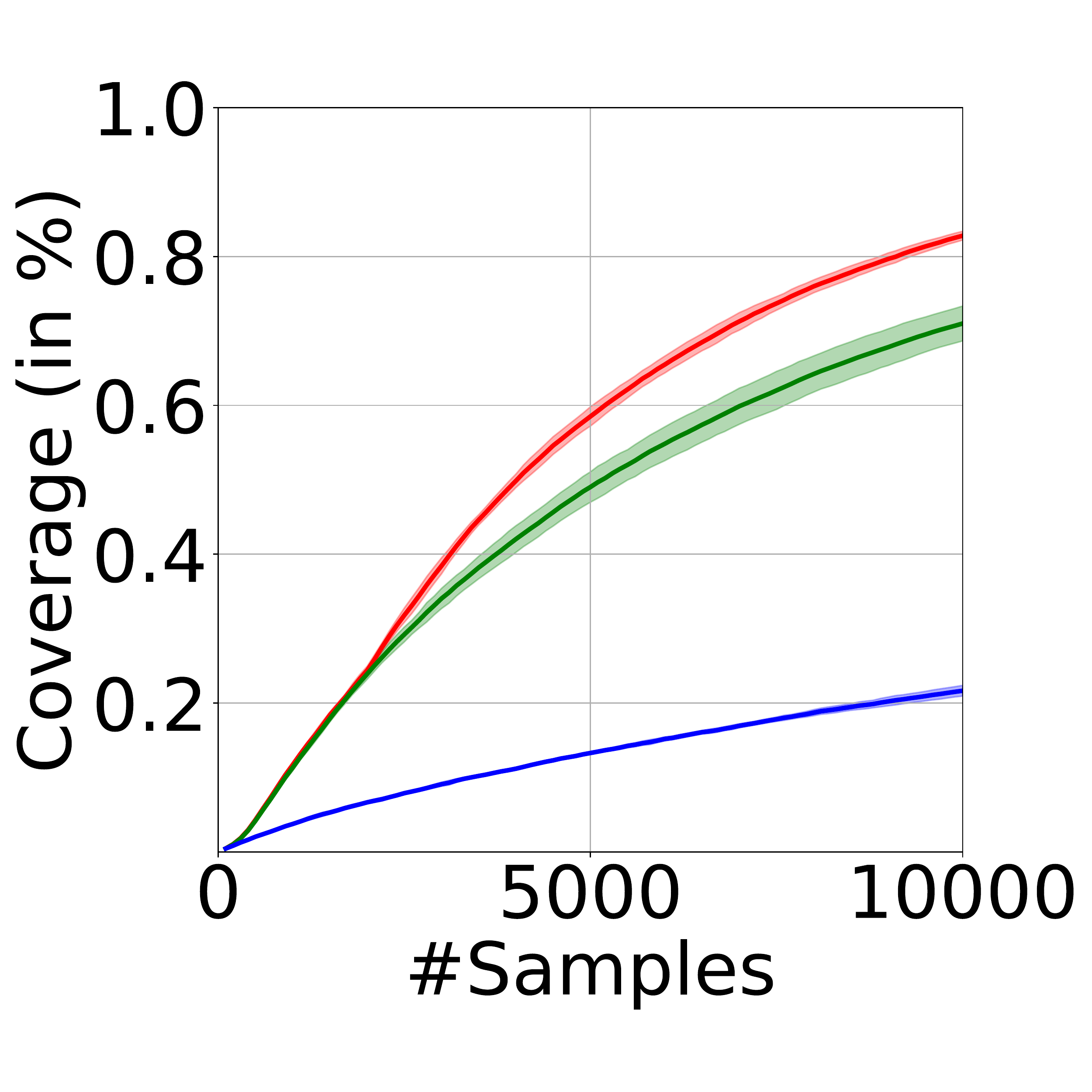}  
  \caption{\texttt{PushCube}}
  \label{figure_vreptask1}
\end{subfigure}%
\begin{subfigure}{.2\textwidth}
  \centering
  \includegraphics[width=.9\linewidth]{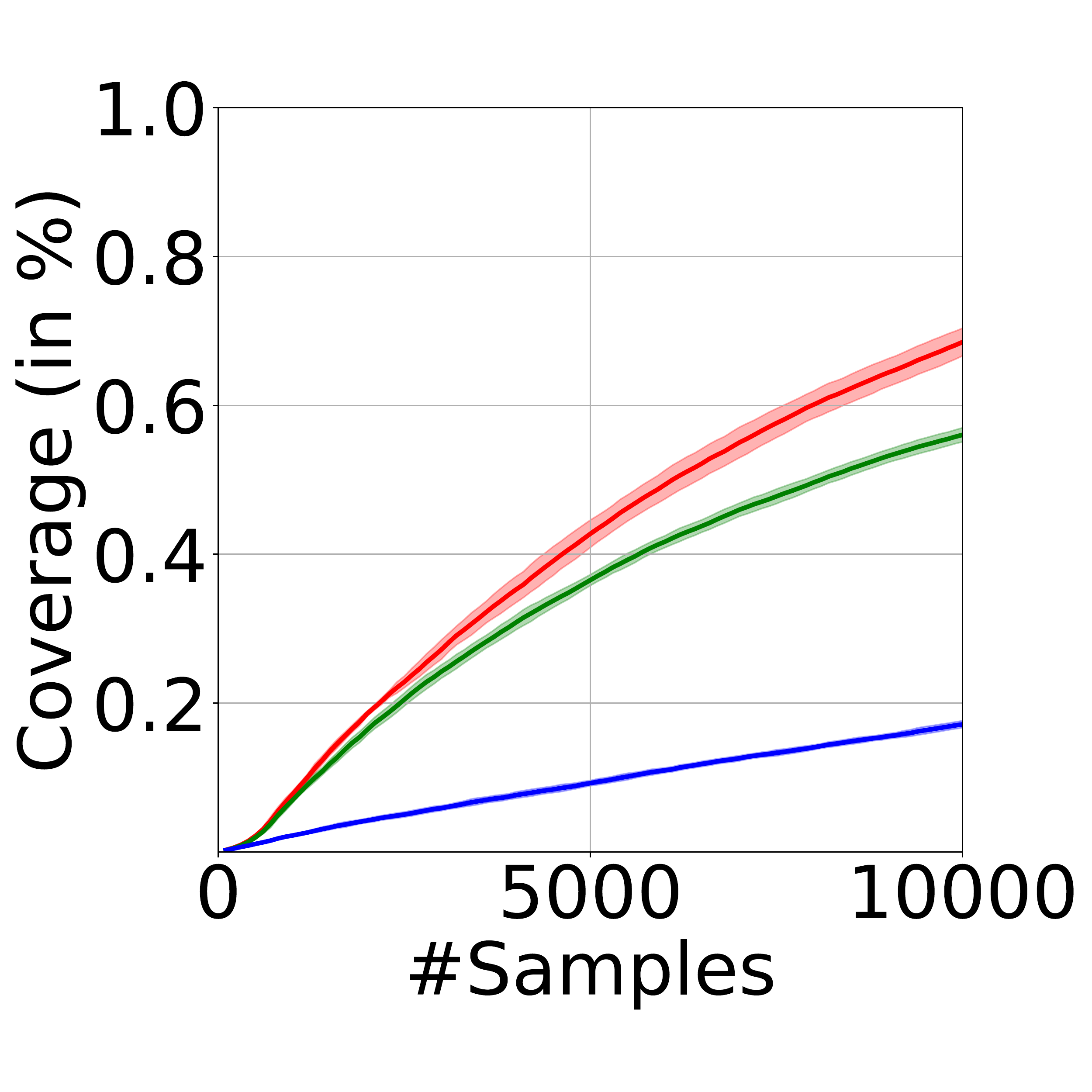}  
  \caption{\texttt{PushSphere}}
  \label{figure_vreptask2}
\end{subfigure}%
\begin{subfigure}{.2\textwidth}
  \centering
  \includegraphics[width=.9\linewidth]{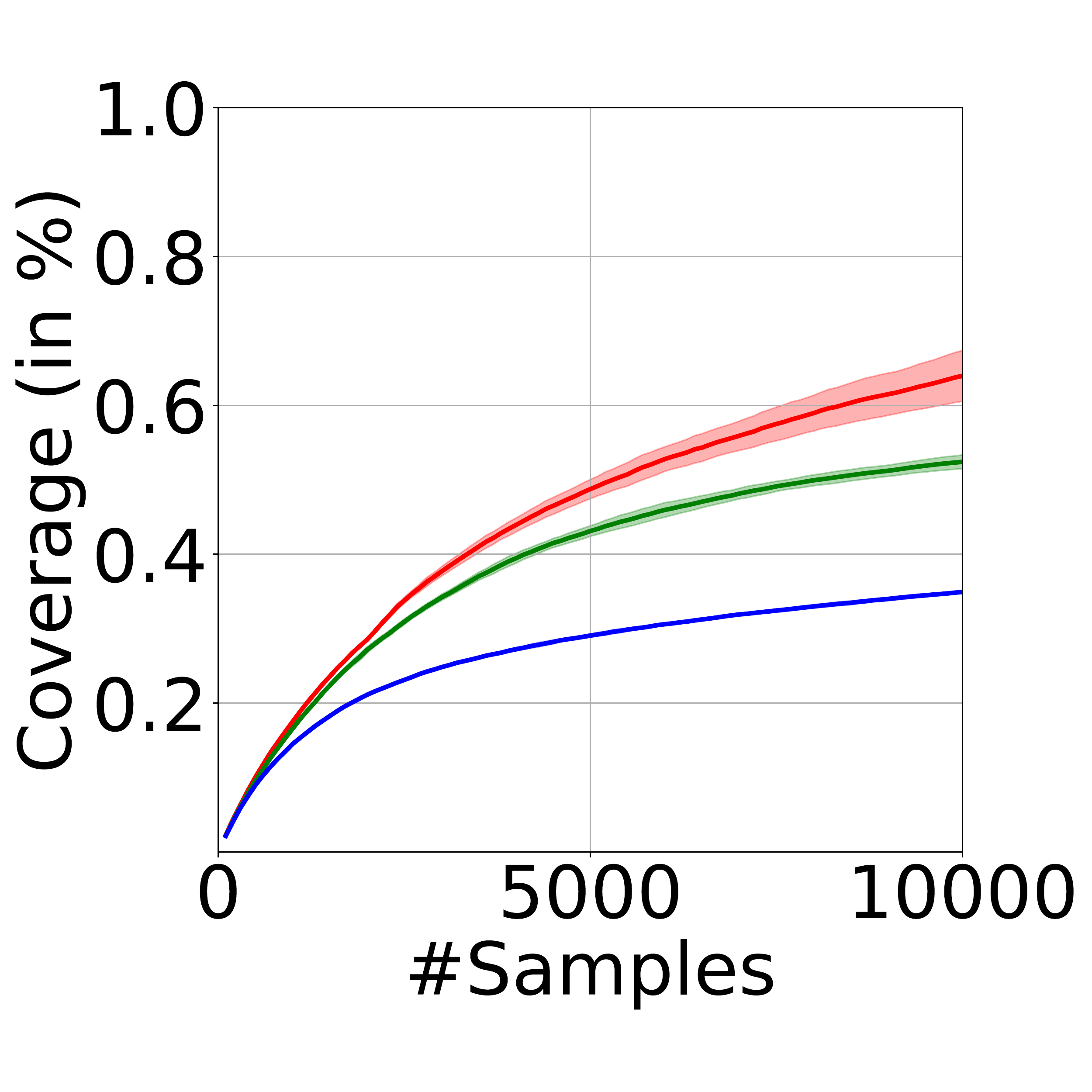}  
  \caption{\texttt{Place}}
  \label{figure_vreptask3}
\end{subfigure}

\caption{(Top) Overlays of the setup of the tasks and repertoires learned using M-QD. Each blue point corresponds to a behavior archived by a skill in the repertoire. (Bottom) Evolution of the behavior-space coverage on the different tasks. Results are averaged over 25 random seeds (10 for the robotic tasks), the plots show the mean and one standard deviation.}
\label{fig_cov}
\vspace{-15pt}
\end{figure*}

\subsection{Model-Based Quality-Diversity Search (M-QD)}
In order to increase the sample-efficiency of the QD-search, we incorporate a model into the evolutionary process described in the previous section.
The model $\phi(a) = (b,q)$ is trained to predict the behavior $b$ and quality $q$ for a given action $a$.
As a model, fully-connected feed-forward neural networks with one hidden layer are used with 
\texttt{ReLU} activation for the neurons in the hidden layer and \texttt{tanh} for the neurons of the output layer, in order to bound the output of the network.
The model is trained in an online fashion by using normalized mini-batches of skills sampled from the repertoire to perform gradient descent steps with the $Adam$~\cite{adam} optimizer after every generation to minimize the mean-squared error.
In order to avoid overfitting to early samples, training only starts after a warm up phase, in which the repertoire is filled with initial skills using the previously described quality-diversity search without the model.

In the evolutionary process, the model is used to predict the behavior $\Tilde{b}$ and the quality $\Tilde{q}$ of all new candidate actions $a$ in the current generation before executing them. Based on these predictions, a novelty and a quality improvement score are estimated for each action
$$
\widetilde{\text{nov}}(a, \Tilde{b}) = \frac{1}{K} \sum_{i=1}^K d(f_b(a_{nn_i}), \Tilde{b}) \ ,
$$
$$
\widetilde{\text{qua}}(a, \Tilde{q}) = \frac{1}{K} \sum_{i=1}^K (\Tilde{q} - f_q(a_{nn_i})) \ ,
$$
where $d(.)$ denotes the Euclidean distance and $\{a_{nn_1}, \dots, a_{nn_K}\}$ are the $K$ nearest neighbours of $a$, measured in behavior space. 
All actions with a novelty score greater than threshold $t_{\text{nov}}$ are evaluated and eventually added to the repertoire, following the same procedure as in the quality-diversity search without model. 
Actions with a predicted novelty below the threshold $t_{\text{nov}}$ are only evaluated if their predicted quality improvement score is greater than the threshold $t_{\text{qua}}$. 
Otherwise, they are discarded without executing them, assuming that they are unpromising and would not improve the repertoire.
Due to this procedure, the sample efficiency is increased as samples that are unlikely to produce a novel behavior or improve a known behavior do not need to be executed in an expensive rollout.

In practice, the hyper parameters $t_{\text{nov}}$ and $t_{\text{qua}}$ need to be chosen carefully.
When chosen too low, no actions would be discarded, nullifying the benefit of the model. 
In contrast, when chosen too high, too many actions would be discarded, putting an excessive confidence on the model's predictions and, thus, hinder exploration of the behavior space.

\subsection{Adaptation of Skills}
While the QD-search typically produces a repertoire which covers the behavior space reasonably well, it is still discrete and thus we have to adapt skills in order to reach arbitrary given behaviors. 
For that purpose, we build upon the gradient descent approach proposed in~\cite{expotocontrol} and extend it with the learned model.
Similar as in~\cite{expotocontrol}, for a given desired behavior $b^*$, the skill which is closest in behavior space $s_{nn} = (a_{nn}, b_{nn}, q_{nn})$ is selected from the repertoire and used to initialize the gradient descent $(a_j, b_j)|_{j=0}=(a_{nn}, b_{nn})$.
Afterwards, a new action is computed by performing one step of the gradient descent given by
\begin{displaymath}
a_{j+1} = a_j + \lambda \widetilde{J}(a_j)^+ (b^* - b_j) \ ,
\end{displaymath}
where $\lambda$ is the adaptation step size.
Evaluating this new action $a_{j+1}$ obtains its behavior $b_{j+1} = f_{b}(a_{j+1})$. 
This process is repeated until the difference between $b_j$ and $b^*$ is smaller than a threshold $t_{\text{dist}}$ or the number of total performed steps is greater than a predefined value.

In~\cite{expotocontrol} the gradient $\widetilde{J}(a_j)$ is estimated from samples in the repertoire  using a least squares approach and a local linearization of the mapping between action and behaviors
$$
\widetilde{J}(a_j) = BG^T(GG^T)^{-1} \ ,
$$
where $G=[a_{nn_1}, \dots, a_{nn_K}]-a_j$ and $B=[b_{nn_1}, \dots,b_{nn_K}]-b_j$ and $\{(a_{nn_1}, b_{nn_1}), \dots, (a_{nn_K}, b_{nn_K})\}$ are the K nearest neighbours of $a_j$ in action space.

In order to obtain a better estimation our adaptation strategy, in contrast, estimates the gradient $\widetilde{J}(a_j)$ using the learned neural network model. 

As our neural network model is differentiable, we can directly use its analytical derivative to obtain the gradient $\widetilde{J}(a_j)$. 
However, given a non differentiable model, the gradient can be estimated, for example, by using a finite difference approach given by
$
\widetilde{J}(a_j) = \frac{\phi(a_j+h) - \phi(a_j)}{h} \ ,
$
where $\phi(.)$ denotes the learned model.

\section{Experiments}
We evaluate and compare the performance of M-QD in two environments and, in total, five different tasks. First, we introduce these tasks and describe their basic setup and goal. 
Consecutively, we report the empirical results of the conducted experiments, demonstrating the benefits of M-QD. 

\subsection{2D Environment}
We first implemented a 2D environment which consists of a bounded two-dimensional space and an agent which is placed on a predefined position in this space. 
An action $a = (a_0, a_1, \dots, a_n)$ is composed of multiple sub-actions $a_i = (\dot{x}, \dot{y}, T)$, which are executed consecutively. 
Each sub-action consists of a velocity in $\dot{x}$ and $\dot{y}$ direction and a duration $T$.
When executing a sub-action, $\dot{x}$ and $\dot{y}$ are directly applied to the agent for $T$ time steps.

\subsubsection{2D Obstacle Task}
For the first task, a number of obstacles are added to the task space. 
An action consists of three sub-actions and the behavior of an action is defined as the final position of the agent.
The goal is to learn a repertoire of skills which enables the agent to move to every point in the space, thus navigating around the obstacles. 
Figure~\ref{figure_toytask1} shows the whole setup of this \texttt{2D Obstacle} task, including the behavior space coverage after learning.

\subsubsection{2D Object Task}
For the second task, a move-able object is placed on a predefined position in the task space. 
An action consists of five sub-actions and the behavior of an action is defined as the final position of this object.
The goal is to learn a repertoire which enables the agent to push this object to an arbitrary position in the space. 
Figure~\ref{figure_toytask2} shows the setup of this \texttt{2D Object} task, including the behavior space coverage after learning.

Formally, for both tasks, the behavior and the quality of an action are defined as 
$$
f_{\text{b}}(a) = (x_{\text{fin}}, y_{\text{fin}}) \text{ and } f_{\text{q}}(a) = \frac{\delta_{\min}}{\delta} \ ,
$$
respectively, where $(x_{\text{fin}}, y_{\text{fin}})$ refers to the final position of the agent or object. The parameter $\delta_{\min}$ refers to the minimum distance the agent/object has to move to reach the behavior corresponding to action $a$, and $\delta$ refers to the actually moved distance of the agent/object. 
Intuitively, if two actions lead to the same behavior, the action which moves a smaller distance is considered to be better.

\subsection{Robotic Environment}
For the second environment, we implemented a more complex robotic environment: It consists of an LBR iiwa 14 R820 robot by KUKA which is placed on a small platform beside a table (Figure~\ref{figure_intro}). 
The joint angles of the robot are preset such that the endeffector points downwards and is exactly placed above the center of the table. 
An action is compounded of multiple via-points $a = (a_1, a_2, \dots, a_n)$, where each via-point $a_i = (x, y, z, \gamma, T)$ consists of an $x$, $y$ and $z$ position as well as an orientation $\gamma$ around the z-axis, and a duration of $T$ time steps.
The $x$- and $y$-orientation are kept fixed, such that the endeffector always points downwards.
When executing an action, the robot moves along these via-points for the given durations.

The Virtual Robotics Experimentation Platform (V-REP) is used for simulation. 
We used the PyRep library~\cite{james2019pyrep}, as it provides faster communication and parallel simulation. 
It is a toolkit for robot learning research, built on top of V-REP.

\subsubsection{Object Pushing Task}
For the first robotic task, a sphere or cube is placed on the center of the table. 
An action consists of two via-points, where the $z$-coordinates are kept fixed to the $z$-coordinate of the sphere or cube to increase the likelihood of interacting with the object. 
The gripper of the robot is kept closed during the whole movement. 
The behavior of an action is defined as the final object position.
The goal is to learn a repertoire which enables the robot to push the object to an arbitrary position on the table.
Figure~\ref{figure_vreptask1} and \ref{figure_vreptask2} show the \texttt{PushCube} and \texttt{PushSphere} task setup, including the coverage of the learned skill repertoires. 
The quality of an action is defined s.t. actions which move the end-effector least are considered best, given by
$$
f_{\text{q}}(a) = - \delta_{ee} \ ,
$$
with $ \delta_{ee}$ denoting the total distance the end-effector moved, measured by the Euclidean distance.

\subsubsection{Object Placing Task}
For the second robotic task, the robot has to learn to place an object safely onto a table and shelves.
Therefore, additionally shelves with different heights are added to the table. 
The robot starts with a cube grasped with its end-effector. 
An action consists of five via-points, which are executed after each other, as described above. 
After the last via-point is reached, the gripper is opened and the cube is released. 
The behavior of an action is defined as the final position of the cube on the table or shelves.
The quality of an action is defined such that skills in which the cube drops as least as possible are considered best, i.e., when the cube is placed safely onto the surface
$$
f_{\text{q}}(a) = d(\text{obj}_s, \text{obj}_e) \ ,
$$
where $\text{obj}_s$ refers to the position of the object after the last via-point is reached but before the gripper is opened, and $\text{obj}_e$ refers to the final position of the cube on the surface. 
The goal is to learn a repertoire which enables the robot to place the cube on every position on the table as safe as possible, whereas safe here refers to a minimal release height. 
Figure~\ref{figure_vreptask2} shows the whole setup and the covered behavior space, we refer to this task as \texttt{Place}. Figure~\ref{figure_intro} illustrates four skills learned on this task.

Formally, for all three robotic tasks the behavior of an action is defined as 
$$
f_{\text{b}}(a) = (x_{\text{fin}}, y_{\text{fin}}) \ ,
$$
where $(x_{\text{fin}}, y_{\text{fin}})$ refers to the final position of the object.

\begin{table}
\renewcommand{\arraystretch}{1.3}
\caption{Final behavior-space coverage ratio of the repertoires. Results are averaged over 10 random seeds (25 for 2D environments), denoted are the mean and one standard deviation. 
}
\label{table_cov}
\centering
\begin{tabular}{c||c|c|c}
\hline
& \bfseries M-QD & \bfseries QD & \bfseries RANDOM\\
\hline\hline
\bfseries \texttt{2D Obstacle} & \bf{0.971 $\pm$ 0.018} & 0.875 $\pm$ 0.017 & 0.731 $\pm$ 0.004\\
\bfseries \texttt{2D Object} & \bf{0.504 $\pm$ 0.073} & 0.366 $\pm$ 0.064 & 0.035 $\pm$ 0.002\\
\bfseries \texttt{PushCube} & \bf{0.828 $\pm$ 0.059} & 0.709 $\pm$ 0.047 & 0.216 $\pm$ 0.073\\
\bfseries \texttt{PushSphere} & \bf{0.685 $\pm$ 0.0185} & 0.560 $\pm$ 0.009 & 0.171 $\pm$ 0.004\\
\bfseries \texttt{Place} & \bf{0.639 $\pm$ 0.034} & 0.524 $\pm$ 0.009 & 0.349 $\pm$ 0.001\\
\hline
\end{tabular}
\vspace{-10pt}
\end{table}

\begin{figure*}[h!]
\begin{minipage}[t]{0.48\textwidth}
\centering
\includegraphics[width=.9\linewidth]{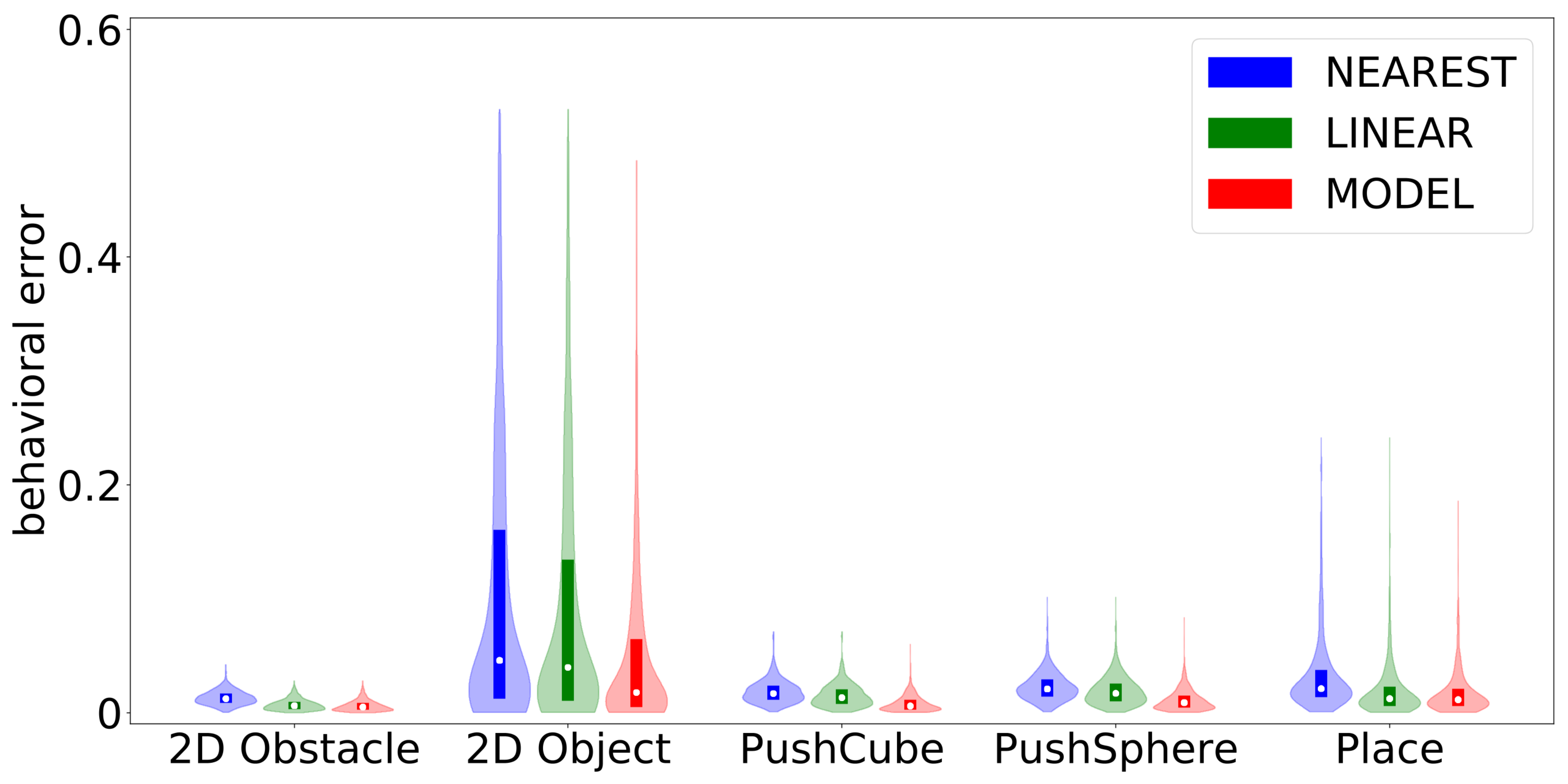}  
  \caption{Behavioral error after applying the different adaptation strategies. 
  Results are averaged over 25 repertoires (10 for the robotic tasks) and 1000 randomly generated desired behaviors for each repertoire. 
  }
  \label{figure_adapt_err}
\end{minipage}\hfill
\begin{minipage}[t]{0.48\textwidth}
\centering
  \includegraphics[width=.9\linewidth]{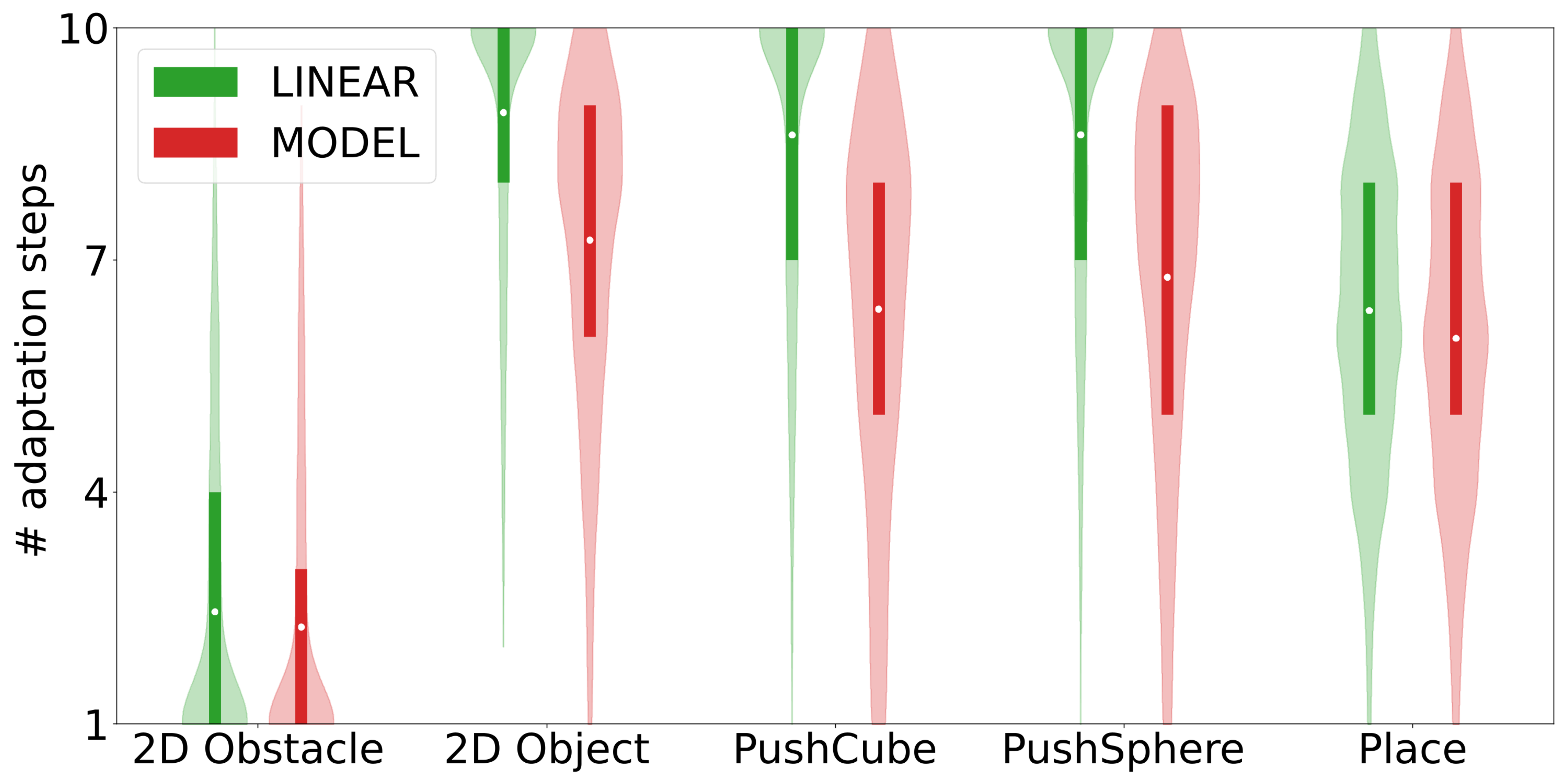}  
  \caption{Number of adaptation steps performed for a behavioral error below the distance threshold, with a maximum of 10 steps. 
  Results are averaged over 25 repertoires (10 for the robotic tasks) and 1000 randomly generated desired behaviors for each. 
  }
  \label{figure_adapt_step}
\end{minipage}
\label{fig:fig}
\vspace{-15pt}
\end{figure*}

\begin{table}[b]
\renewcommand{\arraystretch}{1.0}
\caption{Hyperparameters used in the Experiments}
\label{table_hyper}
\centering
\begin{tabular}{c||c|c}
\hline
& \bfseries 2D & \bfseries Robotic \\
\hline\hline
\bfseries \texttt{Max. generations} & 250 & 100\\
\bfseries \texttt{Population size} &  100 & 100\\
\bfseries \texttt{K} & 5 & 5\\ 
\bfseries \texttt{$t_{\text{dist}}$} & 0.02 & 0.02\\
\bfseries \texttt{$t_{\text{nov}}$} & 0.04 & 0.04\\
\bfseries \texttt{$t_{\text{qua}}$} & 0.0 & 0.0\\
\bfseries \texttt{Learning rate} & 1e-3 & 1e-3\\
\bfseries \texttt{Batch size} & 16 & 16\\
\bfseries \texttt{No. of batches} & 500 & 500\\
\bfseries \texttt{No. of warmup samples} & 1000 & 1000\\
\bfseries \texttt{No. of hidden neurons} & 32 & 64\\
\bfseries \texttt{Adaptation step size} & 0.1 & 0.1\\
\bfseries \texttt{No. of adaptation steps} & 10 & 10\\
\hline
\end{tabular}
\end{table}

\subsection{Results}
In order to evaluate our approach we performed experiments on all five tasks and compared it against two baselines. 

\subsubsection{Building A Repertoire}
We first evaluated the learning of the skill repertoire, and compared the quality-diversity search with and without our proposed model extension. 
Additionally, we compared to another baseline, which we refer to as RANDOM:
This baseline accumulates a repertoire by sampling actions uniformly from the action space. 
We ran all three algorithms for $250$ generations in the 2D Environments and averaged the results over $25$ random seeds. 
For the robotic environments, we ran the algorithms for $100$ generations and $10$ different random seeds. 
We compared the algorithms in terms of average quality and coverage of the behavior space of the final skill repertoire.
In all our tasks, the behavior space is bounded. 
In order to compute the coverage of this bounded space, we define that each skill covers a circular space around its behavior, where the radius of that circle is given by the distance threshold of the quality-diversity search. 
The coverage of a repertoire in percent of the total behavior space is then given by the area of the union of all these circles, divided by the total area of the behavior space. 
The average quality of a repertoire is defined as the average quality of its skills.

Figure~\ref{fig_cov} shows the evolution of the coverage as well as the final repertoires overlayed into the setup of the different tasks.
Table~\ref{table_cov} reports the coverage of the final repertoires.
In terms of coverage, the random search performed poorly on all the tasks compared to the quality-diversity approaches. 
Only on the 2D-Obstacle and the Place task, random search was able to build a repertoire which covers large parts of the behavior space, highlighting the simplicity of the search space in these tasks.
The repertoires of QD and M-QD both cover the behavior space reasonably well, however, M-QD outperforms QD by a margin of more than $10$ percent on all of the tasks, and reaches this coverage with less samples. 
In terms of average quality, M-QD and QD outperformed RANDOM, while the average quality of repertoires generated with M-QD was slightly below the repertoires generated with QD. 
For example, on the \texttt{2D Obstacle} task, RANDOM achieved an average quality of $0.63 \pm 0.02$, while the repertoires generated with QD and M-QD achieved an average quality of $0.83 \pm 0.01$ and $0.80 \pm 0.01$ respectively. 
We observed this small drop in average quality between M-QD and QD on every task, however, we consider it to be negligible, as in our application scenario a good coverage of the behavior space with less samples is more important.

\subsubsection{Adaptation Of Skills} 
Next, the proposed model-based adaptation approach is evaluated.
Therefore, we evaluated the gradient descent approach using the model to estimate the gradient as well as the approach which uses a local linearization.
Additionally we compared the results to just using the nearest neighbour in the repertoire instead of applying an adaptation strategy. 
As repertoire we used the final repertoires produced by the model-based quality-diversity search algorithm for all the strategies. 
We randomly generated $1000$ goals uniformly in behavior space and averaged the results of the different approaches over $25$ different repertoires ($10$ for the robotic tasks).

We report the behavioral error as well as the number of required adaptation steps to achieve these errors in Figure~\ref{figure_adapt_err} \& \ref{figure_adapt_step}. 
Both adaptation strategies outperform nearest neighbour on all the tasks. 
Adaptation with the learned model outperforms the local linearization adaptation. 
The difference between the model-based approach and the local linearization is bigger on more complicated environments and negligible on the simpler ones, as for example \texttt{2D Obstacle} and \texttt{Place}. We suspect that this is the case as in these simple environments the local linearization already yields a reasonable estimate of the true gradient.
Furthermore, the model-based approach requires less adaptation steps than the linear, improving again the sample-efficiency.

\section{Conclusion}

We proposed a Model-based Quality-Diversity search algorithm (M-QD) for learning a diverse repertoire of skills in an open-ended setting.
The approach is evaluated empirically on a 2D environment as well as on a more complex robotic environment. 
The results of our experiments show that M-QD outperforms the model-free quality-diversity search in terms of sample-efficiency and behavior space coverage when learning a skill repertoire. 
Moreover, the results show that in comparison to a simple local linearization approach, using a model-based adaptation strategy reduces the number of needed adaptation steps and the average error in behavior space -- improving the repertoire generalization and skill adaptation efficiency.

However, despite this twofold improved efficiency, the amount of rollouts required by the model-based quality-diversity search algorithm makes learning from scratch on a real system still challenging. 
Thus, in future research, we aim at integrating Sim2Real techniques, as for example presented in~\cite{expotocontrol}, into our approach in order to adapt repertoires and models learned in simulation to real robots, or to concurrently learn in simulation and on the real system. 







\bibliographystyle{IEEEtran}
\bibliography{root}

\end{document}